\documentclass[letterpaper, 10 pt, conference]{ieeeconf}  
\usepackage{multirow}
\usepackage{amssymb,graphicx,amsmath,color,algorithm,algorithmic}
\usepackage[noadjust]{cite} 
\usepackage{caption}
\usepackage{subcaption}
\usepackage{eucal}
\usepackage{layouts}
\usepackage{todonotes}
\usepackage{tabularx,booktabs}
\usepackage{lipsum}
\usepackage{pifont}
\usepackage{pgfplots} 
\usepackage{amssymb}
\newcommand{\cmark}{\ding{51}}%
\newcommand{\xmark}{\ding{55}}%
\usepackage{balance}
\usepackage{paralist}
\usepackage[hidelinks]{hyperref}

\newcolumntype{L}[1]{>{\raggedright\let\newline\\\arraybackslash\hspace{0pt}}m{#1}}
\newcolumntype{C}[1]{>{\centering\let\newline\\\arraybackslash\hspace{0pt}}m{#1}}
\newcolumntype{R}[1]{>{\raggedleft\let\newline\\\arraybackslash\hspace{0pt}}m{#1}}
\pgfplotsset{compat=newest}

\IEEEoverridecommandlockouts                              
\title{\LARGE \bf Learning Whole-body Motor Skills for Humanoids
	\vspace{0em}
}

\author{Chuanyu Yang, Kai Yuan, Wolfgang Merkt, Taku Komura, Sethu Vijayakumar, Zhibin Li
\thanks{All authors are with the Institute of Perception, Action, and Behaviour, School of Informatics, The University of Edinburgh (Informatics Forum, 10 Crichton Street, Edinburgh, EH8 9AB, United Kingdom). Email: {\tt\small {chuanyu.yang@ed.ac.uk}.}}%
}

\begin{document}
	\bstctlcite{IEEEexample:BSTcontrol}
	\maketitle
	\thispagestyle{empty}
	\pagestyle{empty}
	
	\begin{abstract}
		This paper presents a hierarchical framework for Deep Reinforcement Learning that acquires motor skills for a variety of push recovery and balancing behaviors, i.e., ankle, hip, foot tilting, and stepping strategies. The policy is trained in a physics simulator with realistic setting of robot model and low-level impedance control that are easy to transfer the learned skills to real robots. The advantage over traditional methods is the integration of high-level planner and feedback control all in one single coherent policy network, which is generic for learning versatile balancing and recovery motions against unknown perturbations at arbitrary locations (e.g., legs, torso). Furthermore, the proposed framework allows the policy to be learned quickly by many state-of-the-art learning algorithms. By comparing our learned results to studies of preprogrammed, special-purpose controllers in the literature, self-learned skills are comparable in terms of disturbance rejection but with additional advantages of producing a wide range of adaptive, versatile and robust behaviors.
	\end{abstract}
	
	\section{Introduction}
	\label{sec:1}
	
	Legged robots have great potential for being deployed in environments where wheeled robots are limited, such as obstacle obstructed terrain as well as narrow and elevated surfaces (e.g., stairs). However, in contrast to wheeled or tracked robots, humanoids are intrinsically unstable and require active control to balance due to their limited support area, high center of mass, and limited actuator capabilities. Therefore, the range of possible scenarios in which humanoids can be deployed is mostly limited by the humanoids' ability to maintain balance and deal with disturbances and uncertainties. Balance can therefore be considered as one of the core skills for humanoid robots and locomotion.
	
\begin{figure}[t]
		\centering
		\includegraphics[width=0.95\linewidth, trim = {0.0cm 12.0cm 12cm 0.0cm}, clip]{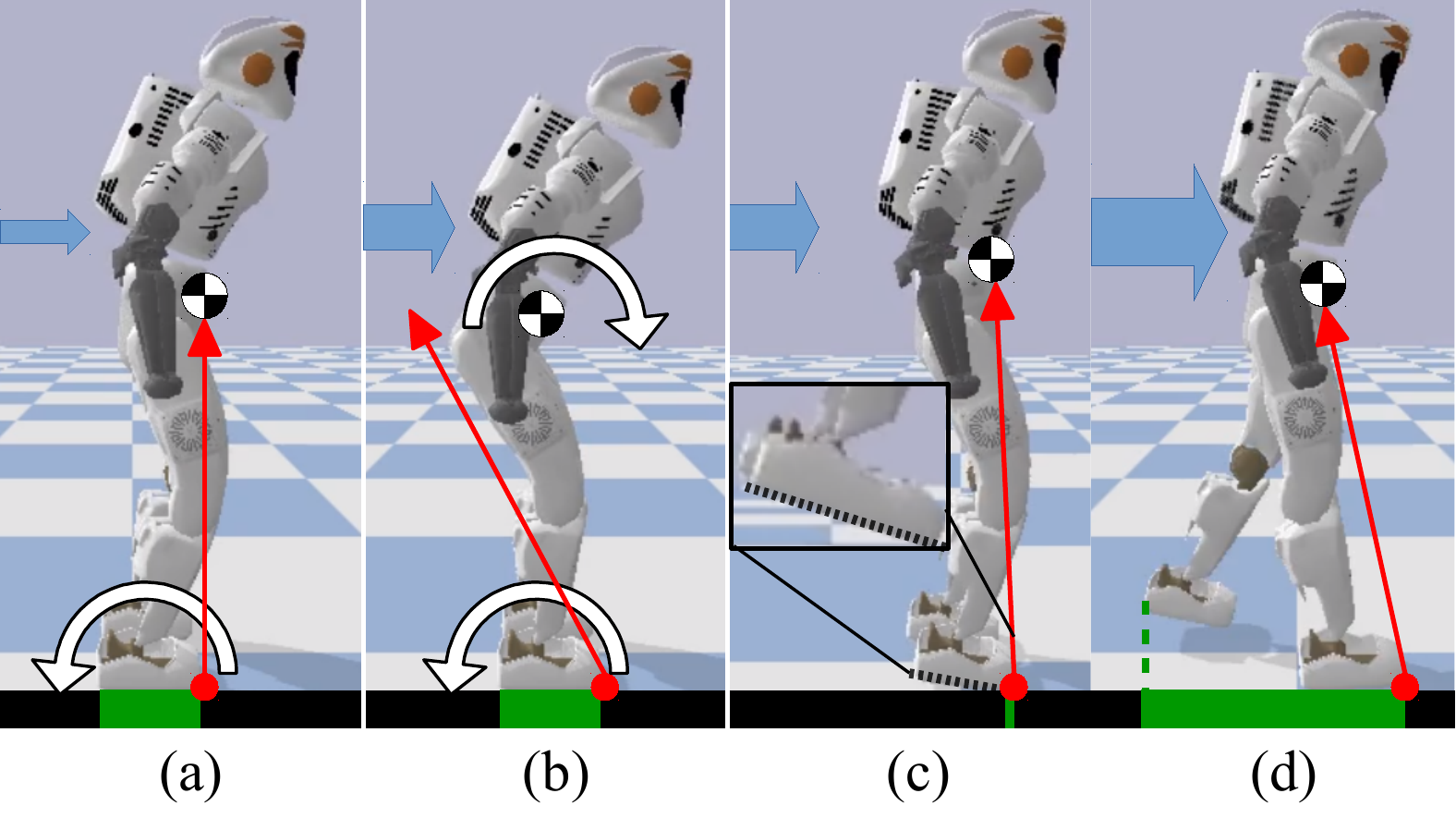}		
		\vspace{0em}
		\caption{Learned push recovery behavior: (a) ankle strategy, (b) hip strategy, (c) foot-tilting strategy, (d) stepping strategy.}
		\label{fig:1}
	\end{figure}

	Classical control methods propose a wide range of balance recovery algorithms, which however lack in the universality of their application. In order to deal with a wide range of pushes, different control strategies need to be applied and traditionally a switching between controllers for the given situation is needed. Generally, different sets of parameters are used for the four main push recovery strategies: ankle, hip, foot-tilting, and stepping (cf. Fig. \ref{fig:1}).
	
	Methods from machine learning on the other hand provide a promising alternative as they can incorporate multiple push recovery skills without the need for hand-tuned gains. 
	Their use is motivated by three main factors as follows.
	
	\begin{figure*}[t]
		\centering
		\begin{subfigure}[t]{.16\textwidth}
			\centering
			\includegraphics[width=\textwidth,trim={0cm 1.2cm 0cm 0.3cm},clip]{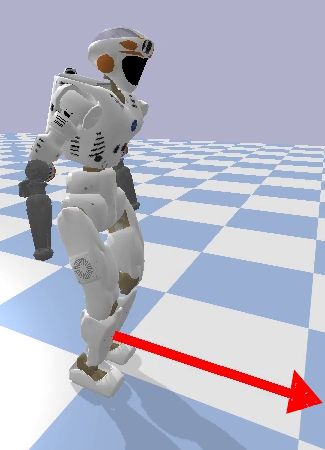}
		\end{subfigure}
		\begin{subfigure}[t]{.16\textwidth}
			\centering
			\includegraphics[width=\textwidth,trim={0cm 1.2cm 0cm 0.3cm},clip]{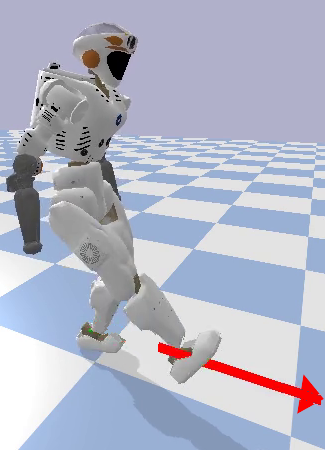}
		\end{subfigure}
		\begin{subfigure}[t]{.16\textwidth}
			\centering
			\includegraphics[width=\textwidth,trim={0cm 1.5cm 0cm 0cm},clip]{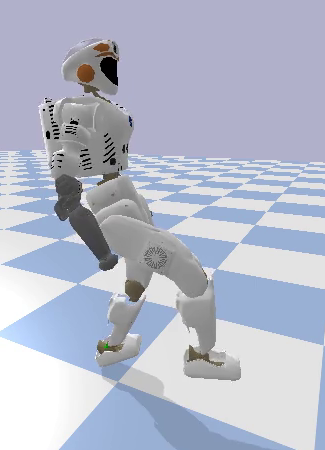}
		\end{subfigure}
		\begin{subfigure}[t]{.16\textwidth}
			\centering
			\includegraphics[width=\textwidth,trim={0cm 1.5cm 0cm 0cm},clip]{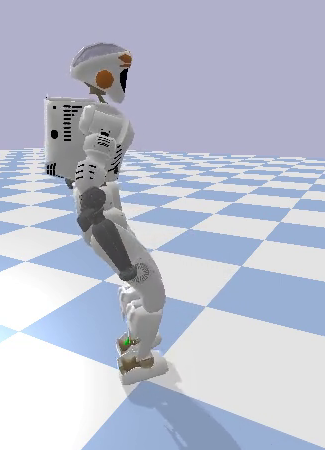}
		\end{subfigure}
		\begin{subfigure}[t]{.16\textwidth}
			\centering
			\includegraphics[width=\textwidth,trim={0cm 1.5cm 0cm 0cm},clip]{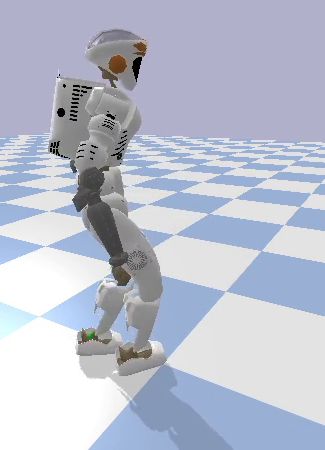}
		\end{subfigure}
		\begin{subfigure}[t]{.16\textwidth}
			\centering
			\includegraphics[width=\textwidth,trim={0cm 1.5cm 0cm 0cm},clip]{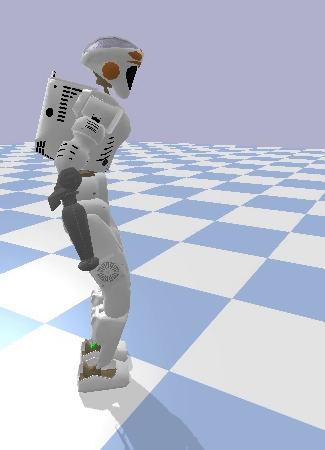}
		\end{subfigure}
		\caption{Snapshots of Valkyrie recovering from an impulse at the shin of $108Ns$, which is a test scenario not encountered during training. The learned policy automatically generates a stepping behavior (cf. \url{https://youtu.be/43ce2cLV0ZI}).}
		\label{fig:snapshots} \vspace{-12pt}
	\end{figure*}
	
	First, supplementing existing control strategies with learning methods allows dealing with scenarios that are hard to engineer in a traditional sense such as sudden, high impact forces and discrete, sudden switches of contact. 

	Second, in contrast to planning and control algorithms \cite{cite:yuan2018AnImprovedFormulation, cite:hu2018ComparisonStudy, cite:li2017RobustFoot} that demand high computational power to run at or close to real-time, e.g. Model-Predictive Control, the computation for machine learning approaches can be outsourced offline. I.e., the computation for Deep Reinforcement Learning (DRL) can be off-loaded into the neural network training phase. By doing so, faster online performance for high dimensional control systems such as humanoids can be achieved. 

	Last, in recent years, DRL has been shown to be capable of solving complex manipulation and locomotion tasks that involve learning a control policy in high-dimensional continuous observation and action spaces~\cite{cite:schulman2015TrustRegion,cite:lillicrap2016ContinuousControl,cite:schulman2017ProximalPolicy}. Instead of manually tuning the control parameters, a feasible policy is learned through interaction with the environment~\cite{dallali2012global}. 

	While there exist various studies using DRL to learn bipedal locomotion for humanoids~\cite{cite:liu2018ImplementationOf,cite:peng2017DeepLoco,cite:yu2018LearningSymmetric}, the used robot models leverage simplified dynamic and collision models and environments in order for faster than real-time simulations at the cost of less realistic simulations. The motivation of this paper is to learn locomotion skills using a realistic robot model obtained from system identification in a realistic simulation environment in order to apply the learned skills on the real robot. We leverage recent advances in DRL to design a unified balance recovery controller which is able to generate sequences of actions that perform similar to or even exceed traditional methods with respect to their disturbance rejection ability. 
	Our work has the following contributions:
	\begin{compactenum}
		\item Application of DRL on an accurate robot model of the Valkyrie platform with realistic settings for simulation. 
		
		\item Design of a learning framework that generates a generic policy. This policy captures a variety of sensor-motor synergies and various control strategies emerge in a unified manner without the need of multiple controllers and the related switching mechanism. 
		
		\item Proposal of a balance recovery specific reward design and training settings of disturbances. These allow the exploration of versatile motions and as a result human-like balancing behaviors, such as foot-tilting and stepping, emerge naturally.
		
		\item Benchmarking the learned policy against control methods. The learned policies generate balance recovery strategies which reject impulses in a similar (or even superior) magnitude as traditionally designed controllers.
	\end{compactenum}
	
	This paper is organized as follows. A brief review on conventional push recovery methods and DRL is presented in Section \ref{sec:2}. Background information on some concepts of DRL and push recovery is explained in Section \ref{sec:3}. The proposed methodology is elaborated in Section \ref{sec:4}. The obtained results are demonstrated and discussed in Section \ref{sec:5}. Finally, a conclusion is drawn in Section \ref{sec:6}.
	
	\section{Related Work}
	\label{sec:2}
	
	\subsection{Conventional Push Recovery methods}
	Over the past two decades, remarkable progress in the field of push recovery for humanoid robots has been made. Without the use of arms, humanoids can leverage four lower body balancing strategies: ankle, hip, stepping, and foot-tilting. The first three strategies, controlling ankle torque, angular momentum around the Center of Mass (CoM), and the timing and position of steps, are analyzed with respect to their ability to reject disturbances in \cite{cite:stephens2007HumanoidPush}. A control framework for the foot-tilting strategy has been proposed in \cite{cite:li2017HumanoidBalancing} demonstrating a humanoid's ability to use foot-tilting for push recovery. Traditionally control schemes can be divided into predictive schemes which calculate reference motions, and reactive schemes which respond to sudden disturbances.
	
	A Model Predictive Control (MPC) scheme that constrains the Center of Pressure to be within the Support Polygon has been proposed in \cite{cite:wieber2006TrajectoryFree}. Strategies involving modulating the Angular Momentum to reactively deal with disturbances have been formally analyzed in \cite{cite:pratt2006CapturePoint,cite:komura2005SimulatingPathological}. Due to the limited size of the contact area, i.e. the foot size, stepping strategies have been proposed \cite{cite:urata2011OnlineDecision,cite:hu2018ComparisonStudy}. This idea was formalized as the Capture Point (CP), the point on which one needs to step in order to come to a complete halt \cite{cite:pratt2006CapturePoint}. Enlarging the support area by stepping has been extended to multi-contact push recovery scenarios in  \cite{cite:han2017FeedbackDesign,cite:marcucci2017ApproximateHybrid}. Methods for balancing on inclined slopes has also been proposed \cite{cite:li2013Stabilizinghumanoids}. Lastly, strategies modulating the height of the CoM in order to compensate for disturbances have been proposed in \cite{cite:koolen2016BalanceControl}. This CoM height modulation can be achieved by either lengthening the leg or in form of foot tilting \cite{cite:li2017HumanoidBalancing}.
	
	\subsection{Deep Reinforcement Learning of Locomotion}
	There exists various successful studies using model-free DRL to solve bipedal locomotion tasks in 3D simulation environments. Schulman et al. proposed a DRL algorithm, Proximal Policy Optimization (PPO), which was applied to successfully learn a locomotion policy that is capable of heading towards a target location in the Roboschool humanoid simulation environment \cite{cite:schulman2017ProximalPolicy}. PPO, together with Deep Deterministic Policy Gradient (DDPG) \cite{cite:lillicrap2016ContinuousControl} and Trust Region Policy Optimization (TRPO) \cite{cite:schulman2015TrustRegion}, are the most commonly used state-of-the-art DRL algorithms for continuous observation-action space control. Further extensions include a parallel computing version of PPO, Distributed Proximal Policy Optimization (DPPO) \cite{cite:heess2017EmergenceOf}, which was applied on a humanoid and successfully learned dynamic and diverse parkour movements for the humanoid character.
	
	Various frameworks have been proposed to allow the DRL agent to learn a policy that generates human-like locomotion behavior for bipedal locomotion tasks.
	Merel et al. proposed a framework that uses generative adversarial imitation learning \cite{cite:ho2016GenerativeAdversarial} to enable the network to learn a policy that produces human-like locomotion gait using limited demonstrations from motion capture data \cite{cite:merel2017LearningHuman}. Peng et al. proposed a framework that incorporated imitation learning by reshaping the reward through the introduction of an imitation term that provides higher reward when the motion is closer to the reference motion capture data \cite{cite:peng2018Deepmimic}. 

	\section{Background}
	\label{sec:3}
	\subsection{Software Setup}
	\label{sec:3a}
	The simulation environment of the Valkyrie robot is built using PyBullet \cite{cite:coumanspybullet} (Fig. \ref{fig:snapshots}). The robot model used in the simulation is the NASA Valkyrie robot \cite{cite:radford2015ValkyrieNasa} with realistic inertia, center of mass, and joint actuation limits. Self-collisions are enabled in the simulation. The DRL algorithm is built using Tensorflow \cite{cite:tensorflow2015-whitepaper}.
	
	\subsection{Deep Reinforcement Learning}
	\label{sec:3b}
	For learning a suitable policy, Deep Reinforcement Learning, particularly model-free policy gradient methods, are used. Policy gradient algorithms operate by maximizing the direct sum of rewards with reference to a stochastic policy. The policy gradient algorithms used in this work are the TRPO \cite{cite:schulman2015TrustRegion}, PPO \cite{cite:schulman2017ProximalPolicy}, and DDPG \cite{cite:lillicrap2016ContinuousControl}. Due to resulting in the best and most robust policy, the TRPO algorithm will be outlined in the following.
	
	\subsubsection{Trust Region Policy Optimization}
	
	In practice, policy gradient methods suffer from high variance which can lead to fluctuations in the performance of the policy between iterations. This problem of instability during training is remedied by introducing a trust region to the numerical optimization which takes a step in the improving direction within a determined trust region. By constraining the amount of changes to the parameters, measured by the Kullback-Leibler (KL) divergence, TRPO guarantees a theoretical monotonic performance improvement of the reward.
	
	For every parameter update iteration, TRPO performs several rollouts and stores the state $s_{t}$, action $a_{t}$ and reward $r_{t}$ into a batch $\mathcal{D}$ until enough data samples are collected, which will then start the update process. During the update process, TRPO updates the policy parameters by minimizing a surrogate loss function while constraining the KL divergence between the new and old policies $\pi_{\theta}$, $\pi_{\theta_{old}}$ to remain within a trust region: 
	\begin{align}
	\min_{\theta} L_{\theta_{old}}(\theta) &= -\mathbb{E}_{t} \left[\frac{\pi_{\theta}(a_{t}|s_{t})}{\pi_{\theta_{old}}(a_{t}|s_{t})}A_{t} \right]\\
	\text{subject to } &\mathbb{E}_{t} \left[KL\left[ \pi_{\theta_{old}}(\cdot|s_{t})\, \pi_{\theta}(\cdot|s_{t})\right] \right]\leq \delta,
	\end{align}
	where $\delta$ is the hyperparameter that determines the trust region, $A_{t} = R_{t} - V(s_{t})$ is the advantage which is calculated by subtracting the return with a baseline. A value estimation $V(s_{t})$ provided by a critic is used as the baseline. 
	
	\subsubsection{Discounted return}
	The total return is used as an evaluation of performance and is determined by calculating the discounted reward, 
	\begin{equation}
	R_{t} = \sum_{l=0}^{T-t}\gamma^{l}r_{t+l}
	\label{eq:discounted_return},
	\end{equation}
	where $T$ is the total number of samples in an episode and $\gamma$ is the discount factor. The half-life of future rewards is used as a reference to decide the value of the discount factor $\gamma$. For balancing, a time horizon between 0.5s and 2s is close. With a frequency of 25Hz, 0.5s equates to 13 time steps. We choose the discount value in a way that the half-life of the future reward occurs at 0.5s, meaning that the accumulated discount factor equates to 0.5 at 13 time step $\gamma^{13} = 0.5$, hence $\gamma \approx0.95$.
	
	\subsubsection{Generalized Advantage Estimation}
	With the policy gradient method and a stochastic policy, we obtain an unbiased estimate of the gradient of the expected total reward, however the estimated policy gradient has high variance. An effective variance reduction scheme for policy gradients called Generalized Advantage Estimator (GAE) was proposed in \cite{cite:schulman2015HighDimensional}. GAE interpolates between a high bias and low bias estimator through the parameter $\lambda \in [0,1]$. One can adjust the bias/variance trade-off by tuning $\lambda$. 
	The GAE for the parameters $\gamma, \lambda$ at time $t$ is:
	\begin{equation}
	\begin{aligned}
	&A_{t}^{GAE(\gamma,\lambda)}:=\sum_{l=0}^{\infty}(\gamma\lambda)^{l}\delta_{t+1}^{V}\\
	&\delta_{t+1}^{V}=r_{t}+V(s_{t+1})-V(s_{t}).
	\end{aligned}
	\end{equation}
	
	\subsection{Capture point}
	\label{sec:3c}
	The Capture point (CP) describes the point on the ground on which the robot should step on in order to come to a complete rest \cite{cite:pratt2006CapturePoint}, and is defined as: 
	\begin{equation}
	x_{\text{CP}}=x_{\text{\tiny{CoM}}}+\dot{x}_{\text{\tiny{CoM}}}\sqrt{\frac{z_{c}}{g}}, 
	\label{eq:cp}
	\end{equation}
	where $x_{\text{CP}}$ is the CP, $x_{\text{\tiny{CoM}}}, \dot{x}_{\text{\tiny{CoM}}}$ is the CoM position and velocity, $z_{c}$ its height, and $g$ the gravitational constant. 
	
	When the CP is within the support polygon, the robot does not need to perform any footstep to maintain balance. Knowing the feet dimensions and therefore the support polygon, the theoretical maximum impulse which can be rejected without foot-stepping can be approximated as follows \cite{cite:pratt2006CapturePoint}:
	\begin{equation}
	J_{\text{reject}}=m\sqrt{\frac{g}{z_{c}}}\Delta_{\text{COP}} ,
	\label{eq:1}
	\end{equation}
	where $\Delta_{\text{COP}}$ is the distance between the CoM and the closest border of the Support Polygon in the direction of the push. For the nominal, upright-standing pose the dimensions of the Support Polygon of Valkyrie is $0.26m \times 0.38m$, the CoM height is at $1.1m$ , the mass of the Valkyrie robot is $137kg$. Equation (\ref{eq:1}) yields an approximate maximal impulse of 53$Ns$ in the sagittal plane and $78Ns$ in the lateral plane for $\Delta_{\text{COP}} = [0.13m, 0.19m]$.

	\section{Methodology}
	\label{sec:4}
	
	
	\subsection{Hierarchical control framework}
	\label{sec:4a}
\begin{figure}[t]
		\centering
		\vspace{0em}
		\includegraphics[width=0.95\linewidth, trim = 4.5cm 9.0cm 5.5cm 6.5cm, clip]{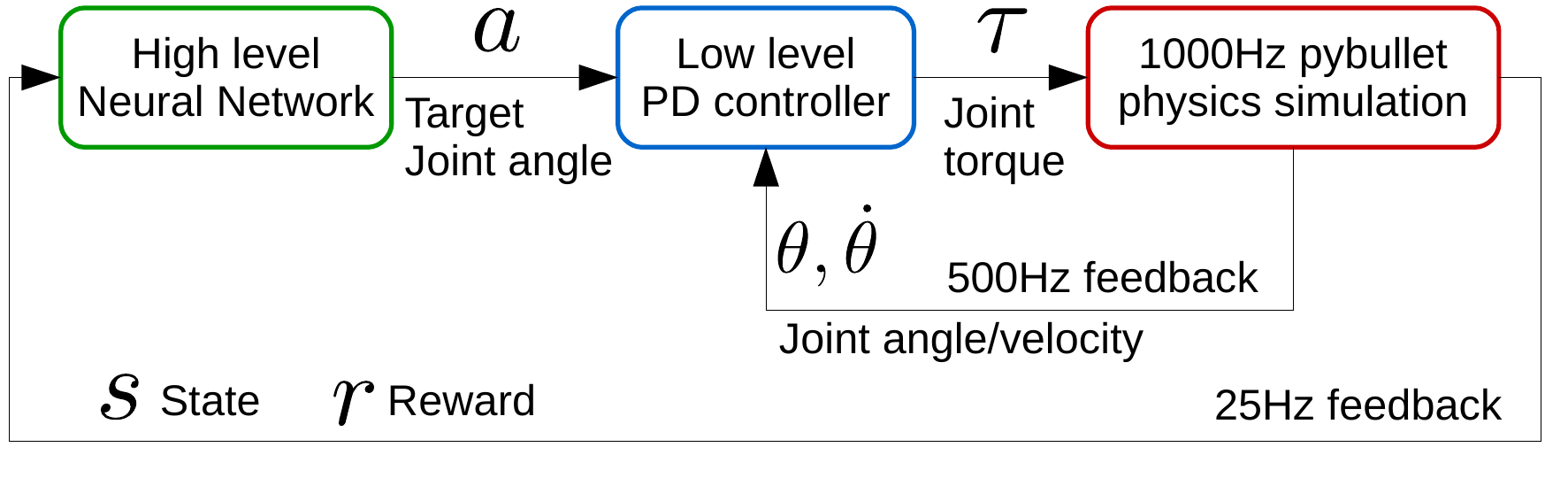}		
		\caption{Hierarchical control system overview \cite{cite:yang2017EmergenceOf}}
		\label{fig:3}
		\vspace{0em}
	\end{figure}

	We designed our control framework to have a hierarchical structure (Fig. \ref{fig:3}). A hierarchical structure allows implementation of two (high and low-level) layers that are independent from each other and can be designed and calibrated separately. The high-level control works under a frequency of 25Hz while the low-level control works at 500Hz. The high-level control is responsible for generating joint angles for a desired motion and the low-level control is responsible for translating the joint angles into joint torques.
	\subsection{Joint-level control of the robot}
	\label{sec:4b}
	Instead of directly controlling the joint motor torque, a PD controller is used to translate joint angles into joint torques. A PD controller resembles the biomechanics of a system in a sense that it has spring damping properties. A comparison between directly using torques and a PD controllers to compute the torques for certain motor tasks was conducted in \cite{cite:peng2016LearningLocomotion} and showed that using a PD controller improved learning speed and overall performance. The resulting torque is computed as:
	\begin{equation}
	u=K_{p}(q_{\text{target}}-q_{\text{measured}})-K_{d}\dot{q}_{\text{measured}},
	\label{11}
	\end{equation}
	where $K_p, K_d$ are the PD gains respectively, $q_{\text{target}}$ is the targeted joint angle, and $q_{\text{measured}}, \dot{q}_{\text{measured}}$ are the measured joint angles and velocities respectively..
	
	\subsection{Observation space and action space}
	\label{sec:4c}
	Input states are chosen in a way such that they can be acquired by sensors on the robot with minimal amount of computation. Immeasurable states are inferred or estimated by the Neural Network. All the sensory information provided as the observational input for the policy is heading-invariant. For balancing, the rotation along the direction of the gravity vector is irrelevant to the balancing state, therefore information about the heading is not needed as feedback, i.e. the policy will perform the same action regardless of global yaw orientation. In order to make the state observation heading invariant, we preprocessed the state by performing transformation of the observations along the gravity axis.
	
	The state $\mathcal{S}\in \mathbb{R}^{47}$ consists of joint angle and velocity, pelvis states (translational and angular velocity, orientation), CoM states (translational velocity and position w.r.t. pelvis), ground contact force, torso position w.r.t. pelvis, and foot position w.r.t pelvis. The observation states are sampled at a frequency of 500Hz and are filtered by a first-order Butterworth filter with a cut-off frequency of 10Hz.
	
	Under consideration of computation efficiency, we minimize the size of action space. A minimum of 11 joints that includes only roll and pitch joints are sufficient for balancing. The action space $\mathcal{A}\in \mathbb{R}^{11}$ of the policy describes the motion of the joint angles. The upper body joints are locked in a nominal position, while for the lower body, only the pitch joint and roll joint are controlled. The controlled joints therefore are: torso pitch, left and right hip pitch \& roll, knee pitch, and ankle pitch \& roll.
	
	
	\subsection{Design of reward function}
	\label{sec:4d}
	The design of the reward function is a crucial part in reinforcement learning as the reward governs the outcome behavior. The reward design follows a similar design rule as in \cite{cite:yang2017EmergenceOf}. Balancing can be divided into four subtasks: regulating upper body pose, regulating CoM position, regulating CoM velocity, and regulating ground contact force. The individual reward is calculated using $\tilde r_i = \exp(-\alpha_{i} (x_{target}-x)^2)$, with $x_{target}$ as the desired value, $x$ as the real value, and $\alpha_{i}$ as the normalization factor. These are then weighted by $w_i$. Furthermore, additional penalty terms are added: ground contact regulation, loss of contact with the ground, and when other parts of the body other than the foot make contact with the ground. We also apply a penalty for the control effort used. The overall reward can be viewed as a sum of the individual reward terms:
	\begin{equation}
	\begin{aligned}
	r  = &r_{pose} + r_{CoM\_pos} + r_{CoM\_vel} + r_{GRF} + \\
	&r_{contact} + r_{power}.
	\label{eq:total_reward}
	\end{aligned}
	\end{equation}
	
	\subsubsection{Upper body pose modulation}
	The upper body pose is represented by the pitch and roll angle of the torso and pelvis. The desired orientation for the pitch roll angle for both pelvis and torso is 0, which is the orientation of the upper body when it is upright:
	\begin{equation}
	\begin{aligned}
	r_{pose} = &w_{\phi_{\text{torsoPitch}}}\tilde r_{\phi_{\text{torsoPitch}}}+ w_{\phi_{\text{pelvisPitch}}}\tilde r_{\phi_{\text{pelvisPitch}}}+\\
	&w_{\phi_{\text{torsoRoll}}}\tilde r_{\phi_{\text{torsoRoll}}}+ w_{\phi_{\text{pelvisRoll}}}\tilde r_{\phi_{\text{pelvisRoll}}}.
	\end{aligned}
	\end{equation}
	
	\subsubsection{CoM position modulation}
	The reward term for CoM modulation is decoupled into horizontal and vertical components. For the horizontal CoM position, the target position is the center of the support polygon to provide maximum disturbance compensation. For the vertical CoM position, the robot should stand upright and maintain a certain height,
	\begin{equation}
	r_{CoM\_pos} = w_{xy_{\text{\tiny{CoM}}}}\tilde r_{xy_{\text{\tiny{CoM}}}} +
	w_{z_{\text{\tiny{CoM}}}}\tilde r_{z_{\text{\tiny{CoM}}}}.
	\end{equation}
	
	\subsubsection{CoM velocity modulation}
	Similar to the CoM position, the reward for CoM velocity is decoupled into two components: velocity in the horizontal and vertical planes. The CoM velocity is represented in the world frame. The desired vertical CoM velocity is 0 as we want to minimize vertical movement, while the desired velocity for horizontal CoM velocity is derived from capture point \eqref{eq:cp}. The desired COM velocity is calculated following the method presented in \cite{cite:yang2017EmergenceOf}. The Capture Point is only valid when the robot has contact with the ground with no slipping, therefore when the robot is in the air, the reward term for horizontal CoM velocity $\tilde r_{\dot{xy}_{\text{\tiny{CoM}}}}$ is deemed invalid and is set to 0:
	\begin{equation}
	r_{CoM\_vel} =  \begin{cases}
	w_{\dot{xy}_{\text{\tiny{CoM}}}}\tilde r_{\dot{xy}_{\text{\tiny{CoM}}}}+
	w_{\dot{z}_{\text{\tiny{CoM}}}}\tilde r_{\dot{z}_{\text{\tiny{CoM}}}}, &\text{foot contact}\\
	w_{\dot{z}_{\text{\tiny{CoM}}}}\tilde r_{\dot{z}_{\text{\tiny{CoM}}}}, &\text{no foot contact}.
	\end{cases}
	\end{equation}
	
	\subsubsection{Contact force modulation}
	The force has to be evenly distributed between both feet for a stable robust balance. The total mass of 137kg yields a force of 671.3N for each foot:
	\begin{equation}
	r_{GRF} = w_{F_{left}}\tilde r_{F_{left}}+w_{F_{right}}\tilde r_{F_{right}}.
	\end{equation}
	
	\subsubsection{Ground contact}
	When the robot is standing, only the feet are in contact with the ground, therefore a penalty is introduced whenever both feet lose contact with the ground or body parts other than the feet make contact with the ground:
	\begin{equation}
	r_{contact} = \begin{cases}
	-2, &\text{if no foot contact with ground}\\
	-10, &\text{if upper body contact with ground}.
	\end{cases}
	\end{equation}
	
	\subsubsection{Power consumption}
	The power consumption is calculated as follows:
	\begin{equation}
	\begin{aligned}
	r_{power}  = w_{power}\cdot\sum_{j=0}^{11}\left \| \tau^{j} \cdot \dot q^{j} \right \|
	,
	\label{eq:3}
	\end{aligned}
	\end{equation}
	 with $\tau^{j}$ is the torque applied on individual joints, and $\dot q^{j}$ is the joint velocity.
	 
	\subsection{Network structure}
	\label{sec:4e}
	The stochastic policy $\pi_{\theta}(a|s)$ is represented as a conditional Gaussian policy $\pi_{\theta}(a|s) \sim \mathcal{N}(\mu_{\theta}(s), \sigma_{\theta})$. The mean of the Gaussian policy is parametrized by a neural network with parameters $\theta$, the covariance of the Gaussian policy is independent from the neural network and is maintained by a separate set of parameters $\sigma_{\theta}$.
	
	The critic $V_{\phi}$ parametrizes the value function with a separate set of neural networks using parameters $\phi$. Both the actor and the critic are parametrized by a fully connected feedforward neural network that consists of 3 hidden layers with 100, 50 and 25 neurons for each layer. The actor network uses $\tanh$ activation for the hidden layers while the critic uses ReLU activation for the hidden layers. The output of both network is produced by linear activation.
	
	The actor network is trained to maximize the reward function (section \ref{sec:4d}), while the critic network is trained by minimizing the loss function $L_{\text{V}}(\phi)$:
	\begin{equation}
	\begin{aligned}
	L_{\text{V}}(\phi) = \mathbb{E}_{t}\left[(V(s_{t})-y_{t})^2\right],
	\end{aligned}
	\label{eq:7}
	\end{equation}
	with the discounted Return $y_t$ (eq. \ref{eq:discounted_return}), value function $V(s_t)$.
	\begin{figure}[t]
		\centering
		\includegraphics[width=0.90\linewidth, trim = {0.0cm 0.75cm 5.0cm 0.0cm}, clip]{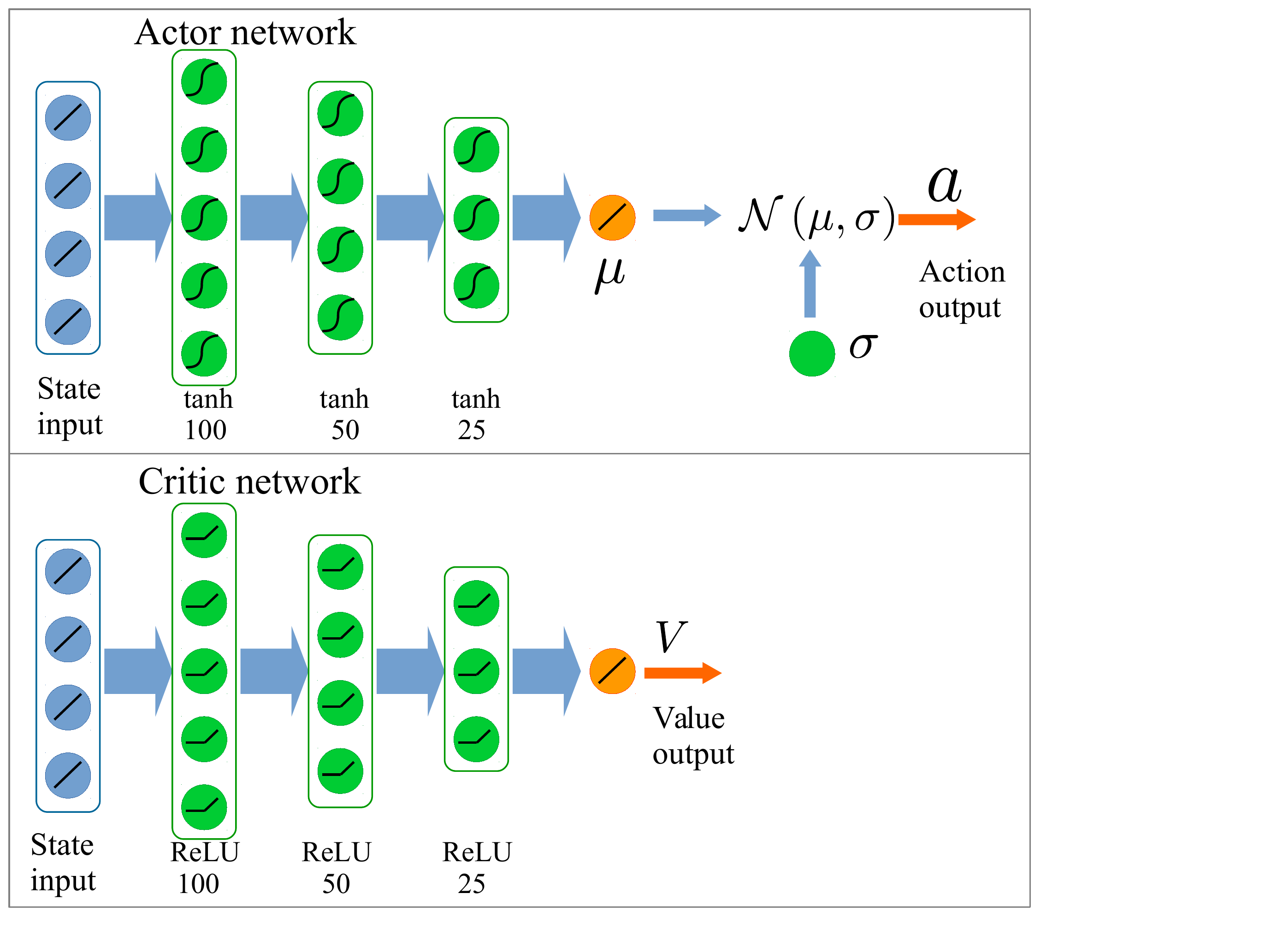}		
		\vspace{0em}
		\caption{Overview of neural network structure.}
		\label{fig:4}
	\end{figure} 
	
	
	\subsection{Exploration during training}
	\label{sec:4f}
	In order to learn a policy capable of withstanding large push disturbances, sufficient exploration during the training phase needs to be provided. Therefore, in addition to the stochastic policy, random forces are applied on the pelvis during the training. From Capture Point theory, the maximum disturbance in the sagittal plane without foot-stepping is 53$Ns$. The bounds of the training disturbances is chosen to be $[53\times0.5Ns, 53\times2Ns]$. The orientation of the force in the horizontal plane and the disturbance in the bound are randomized. Disturbances are applied to the robot multiple times during each trial, with 5s interval between subsequent push disturbances for push recovery.
	
	\subsection{Learning Algorithm}
	\label{sec:4g}
\begin{figure}[t]
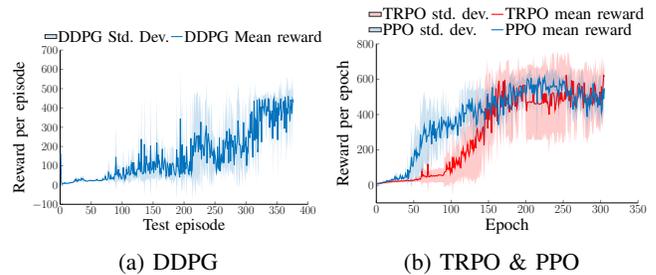

		\centering
		\begin{subfigure}[t]{.235\textwidth}
			\centering
			\resizebox{\textwidth}{!}{\input{ddpg_reward.tex}}
			\caption{DDPG}
		\end{subfigure}
		\begin{subfigure}[t]{.235\textwidth}
			\centering
			\resizebox{\textwidth}{!}{\input{ppo_trpo_reward.tex}}
			\caption{TRPO \& PPO}
		\end{subfigure}
		\caption{Learning curves for DDPG, PPO, and TRPO. The performance are evaluated using the deterministic policy. The mean of the Gaussian policy learned by PPO/TRPO is used for evaluation. The results are averaged over 7 learning trials.}
		\label{fig:ddpg_vs_trpo_ppo}
	\end{figure}

	Due to the structure and choice of our framework, the learned policy is independent of the type of learning algorithm. We trained a policy for maintaining balance via TRPO, PPO, and DDPG, and found similar resulting behavior (cf. Table \ref{tab:comparison_trpo_ppo_ddpg}). All four balancing strategies (Fig. \ref{fig:1}) emerges regardless of the DRL algorithm used. However, from our simulations, TRPO is able to achieve higher rewards and is able to withstand higher impulses. Figure \ref{fig:ddpg_vs_trpo_ppo} shows the learning curves for the policies learned in Table \ref{tab:comparison_trpo_ppo_ddpg}. DDPG is trained off-policy and utilizes a replay buffer, whereas TRPO and PPO are trained on-policy batch-wise, which makes it  difficult to directly compare.
	
	All three DRL algorithms are able to learn a feasible balancing policy. The difference in performance can be attributed to the randomness in different trials of training and hyperparameters. Training is performed entirely on a single Intel Core i7-6700K with 4.0 GHz and converges in two days.
	
	\begin{table}[h!]
		\centering
		\caption{Maximal rejectable impulses for the various learning algorithms without taking steps.}
		\begin{tabular}{L{2.2cm}C{1.6cm}C{1.6cm}}
			\toprule
			&  \multicolumn{2}{c}{Maximal  disturbance in $Ns$}\\ \cmidrule{2-3}
			Learning algorithm	&  Sagittal  & Lateral \\ 
			\cmidrule{1-3}
			TRPO &  240 & 78 \\
			DDPG &  75 & 160 \\
			PPO &   192 & 36 \\
			Baseline from \eqref{eq:1} & 53 & 78\\
			\bottomrule
		\end{tabular}
		\label{tab:comparison_trpo_ppo_ddpg}
	\end{table}
	
	%
	

	\section{Results}
	\label{sec:5}
	In the following, a series of test scenarios are presented to evaluate the performance of the control policy acquired by the deep reinforcement learning agent. Furthermore, we show its robustness to external disturbances, as well as noise in the observation (measurement) and action (actuation) spaces. Next, a comparison against traditional methods from other works is made. Lastly, the physical validity of the generated motions is analysed and verified. Please refer to the accompanying video for the results (cf. \url{https://youtu.be/43ce2cLV0ZI}).
	
	\subsection{Horizontal push on pelvis}
\begin{figure*}[th!]
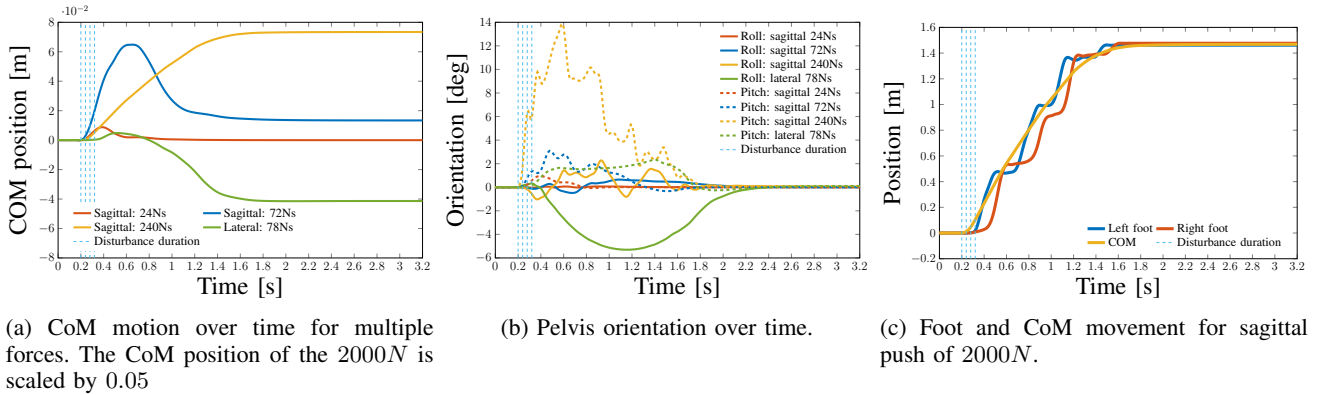

		\centering
		\begin{subfigure}[t]{.32\textwidth}
			\centering
			\resizebox{\textwidth}{!}{\input{horizontal_push_COM.tex}}
			\caption{CoM motion over time for multiple forces. The CoM position of the $2000N$ is scaled by $0.05$}
		\end{subfigure}
		\begin{subfigure}[t]{.32\textwidth}
			\centering
			\resizebox{\textwidth}{!}{\input{horizontal_push_angular_momentum.tex}}
			\caption{Pelvis orientation over time.}
		\end{subfigure}
		\begin{subfigure}[t]{.32\textwidth}
			\centering
			\resizebox{\textwidth}{!}{\input{foot_2000N.tex}}
			\caption{Foot and CoM movement for sagittal push of $2000N$.}
		\end{subfigure}
		\caption{Resulting motions from impulse disturbance and balance recovery. }
		\label{fig:horizontal_push}
	\end{figure*}

	Horizontal disturbances were applied on the pelvis during the training phase, and therefore the DRL agent should be able to learn to withstand such type of disturbances. The robot exhibits different behavior depending on the amount of disturbance applied (Table \ref{tab:horizonal_pushes}, Fig. \ref{fig:horizontal_push}). Different control strategies emerge and range from generating ankle torque to shift the COP (ankle strategy), generating angular momentum (hip strategy), over tilting the foot to dissipate the disturbance (foot tilt strategy), to taking a step to recover from the large push (stepping strategy). The magnitude of the lateral pushes, for which the robot is capable of withstanding, is significantly smaller than in the sagittal plane. This is due to the fact that the support leg will block the swing leg in the lateral direction, limiting the range for leg movement. Dealing with this problem involves either jumping to take a step, or crossing the legs, which, due to the kinematic constraints, is not possible for Valkyrie. These jumping manoeuvres were not learned by the policy, as high velocities in the CoM resulted in lower rewards.
	
	From the CoM position (Fig. \ref{fig:horizontal_push}a) and the pelvis orientation (Fig. \ref{fig:horizontal_push}b) it can be inferred that the robot is standing still after 3s. As can be seen by the eight flat plateaus in Figure \ref{fig:horizontal_push}(c), eight steps are taken in order to deal with a impulse disturbance of $240Ns$ at the pelvis. After the eighth step the robot stands still in the nominal pose. For other horizontal pushes, the stepping behavior is similar.
	\begin{table}[h!]
		\centering
		\caption{Emerging behavior for impulse disturbances of different magnitudes. A checkmark indicates that the respective strategy is applied in addition to the other marked strategies.}
		\begin{tabular}{L{2.5cm}C{0.9cm}C{0.9cm}C{0.9cm}|C{0.9cm}}
			\toprule
			&  \multicolumn{4}{c}{Impulse disturbance in $Ns$}\\ 
			\cmidrule{2-5}
			Emerging behaviour	&  $24Ns$  & $72Ns$  & $240Ns$  & $78Ns$   \\ 
			\cmidrule{1-5}
			Push direction     & sagittal & sagittal & sagittal & lateral \\ 
			Ankle strategy &  \cmark & \cmark & \cmark & \cmark  \\
			Hip strategy&  \cmark & \cmark & \cmark & \cmark  \\
			Foot tilt strategy &   \xmark & \cmark & \cmark & \cmark \\
			Stepping strategy&   \xmark & \xmark & \cmark & \xmark \\
			\bottomrule
		\end{tabular}
		\label{tab:horizonal_pushes}
	\end{table}
	
	\subsection{Force disturbance on other body segments}
	During the training phase only horizontal disturbances were applied on the pelvis. It is well known in machine learning that the test set should vary from the training set. Therefore, we also designed test scenarios which the DRL agent has never encountered before during training to see how well the policy generalizes.
	
\begin{figure*}[th!]
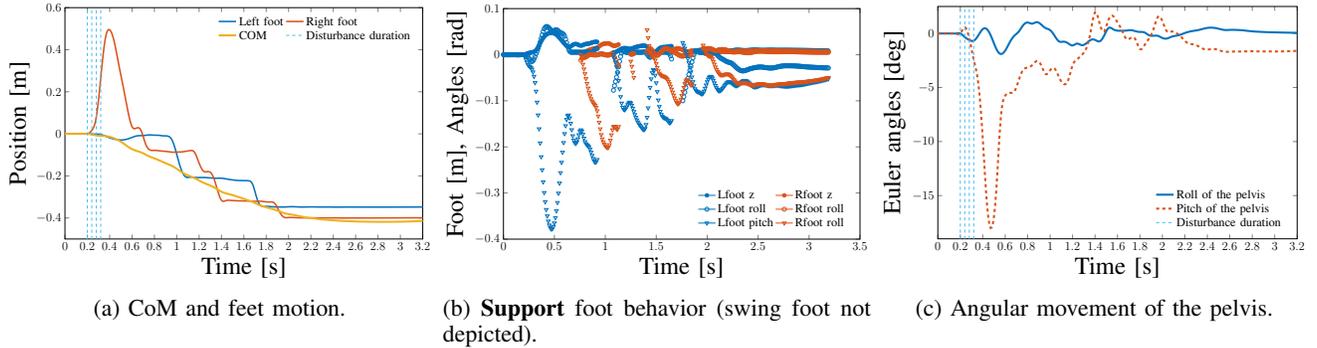

		\centering
		\begin{subfigure}[t]{.32\textwidth}
			\centering
			\resizebox{\textwidth}{!}{\input{thigh_foot_1000N.tex}}
			\caption{CoM and feet motion. }
		\end{subfigure}
		\begin{subfigure}[t]{.32\textwidth}
			\centering
			\resizebox{\textwidth}{!}{\input{thigh_foot_tilt.tex}}
			\caption{\textbf{Support} foot behavior (swing foot not depicted).}
		\end{subfigure}
		\begin{subfigure}[t]{.32\textwidth}
			\centering
			\resizebox{\textwidth}{!}{\input{thigh_angular_momentum.tex}}
			\caption{Angular movement of the pelvis.}
		\end{subfigure}
		\caption{Resulting motions from an impulse disturbance at the shank. The robot takes 6 steps before standing stably. }
		\label{fig:pull_parts}	
	\end{figure*}

	In push recovery studies, the disturbance is usually applied near the CoM to avoid introducing any torque into the system, as it is more challenging to balance a robot with high amount of angular momentum. We are interested in how well the policy will perform when disturbance is applied on other parts of the body far away from the CoM. We chose three body parts for which a large torque and angular movements would result when force is applied on: the upper torso, leg thigh, and leg shank (Table \ref{tab:other_part_pushes}). The resulting motion for being pulled at the shank can be seen in Figure \ref{fig:pull_parts}. In Figure \ref{fig:pull_parts}a) six steps for recovering balance are observed. The support foot height, and roll and pitch angle relative to the ground can be seen in Figure \ref{fig:pull_parts}b). Finally, the angular movement of the pelvis Euler angles (Fig. \ref{fig:pull_parts}c) show that the robot recovers into an almost nominal pose after six steps.
	\begin{table}[h!]
		\centering
		\caption{Maximum rejected impulse for different body parts.}
		\begin{tabular}{L{1.4cm}C{1.8cm}C{1.0cm}C{1.3cm}C{1.0cm}}
			\toprule
			Body part &  Max. impulse in $[Ns]$ & Lever in $[m]$ & Torque in $[Nm]$ & Amount of steps  \\ 
			\cmidrule{1-5}
			Upper torso &  120 & 0.32 &320 &6\\
			Leg thigh &  108 &0.50 & 450& 4\\
			Leg shank &   108 & 0.70 & 630 &6\\
			\bottomrule
		\end{tabular}
		\label{tab:other_part_pushes}
	\end{table}
	
	\subsection{Landing from height}
	A different type of impact is applied to the robot by dropping it from a height above the ground with the objective to land stably. Landing involves high impact and sudden changes in ground contact. Various control strategies involving the Zero-Moment Point (ZMP) or CP assume steady contact with the ground and, without a switching mechanism, fail to perform when the robot is in the air. This test scenario is used to test whether the policy can handle high impact and sudden change in ground contact and land stably.
	
	The robot is capable of landing after being dropped from a maximum height of $0.55m$ above the ground. Furthermore, it is able to handle randomly initialized orientations of the pelvis within $[-5^{\circ},5^{\circ}]$ under a dropping height of 0.4m (with a leg length of $1m$, this is a displacement of $0.087m = 1m\cdot \sin(5\frac{\pi}{180})$ compared to the initial position).
	
	\subsection{Combined test case}
	Lastly, a combined test case involving linear and angular momentum disturbances is designed. In this test scenario, we first apply a vertical force upward to lift the robot off the ground and then apply a horizontal force when the robot is in the air. As a result, the robot has to handle linear and angular momentum and sudden high impacts at the same time. The robot is able to recover for a vertical impulse of $500Ns$, and a horizontal impulse of $70Ns$. In order to deal with this combined disturbance, the policy used all four push recovery strategies: ankle, hip, toe, step. Apart from the physical impossibility of throwing a 137kg robot, the resulting joint torques, velocities are within the joint limits of the real Valkyrie (cf. Section~\ref{sec:reasonable}).
	
	\subsection{Robustness against noise in observation and action space}
	In real world applications, noisy measurements and actuations signals are a main contributor to the discrepancy between simulations and reality. In order to test our policy's ability to be applied in the real world, its ability to handle noise both in action and observation space is tested. For this, a Gaussian distributed noise $d \sim \mathcal{N}(\mu, \sigma)$ is added to both action space ($\mu = 0, \sigma = 0.1$), and state space ($\mu = 0, \sigma = 0.5$). We found that the policy is able to handle both. The observation noise is filtered by the Butterworth filter, while noise in the action space is handled by the robustness of the policy.
	
	\subsection{Comparison against other control methods}
	In order to compare the results obtained from the policy with other works, the disturbance is normalized. The applied force needs to be put in relation with the duration of the push, resulting in the impulse acting on the system. Furthermore, the mass and the resulting inertia is a crucial variable to a robot's ability to deal with disturbances. Therefore, the impulse is normalized by the weight of the robot. 
	
	The normalized impulse is used for comparison between the controllers of other works \cite{cite:urata2011OnlineDecision,cite:wieber2006TrajectoryFree,WIEBER2006559,cite:stephens2010PushRecovery} and the learned policy (Table \ref{tab:1}). 
	We compared sagittal and lateral pushes. For the sagittal push, two impulses are chosen such that foot stepping occurs for the larger impulse ($1.73Ns/kg$), while the smaller impulse ($0.57Ns/kg$) will result in a strategy without stepping. 
	By comparing the rejectable normalized impulse of the strategies not taking a step (A, B, E, G), it can be seen that our policy performs similar (E: $0.57Ns/kg$, G: $0.56Ns/kg$) to the other controllers (A: $0.6Ns/kg$, B: $0.52Ns/kg$). For the strategies taking a step (C, D, F), our policy is able to perform better (F: $1.73Ns/kg$) than the stepping controllers C ($0.52Ns/kg$) and D ($0.6Ns/kg$). Albeit our results are obtained from simulation whereas C and D are obtained from real experiments, we show in the next section that the generated motions are realistic and within the real physical constraints. 
	
	\begin{table*}[t]
		\centering
		\caption{Push disturbance from various push recovery studies}
		\label{tab:1}	  
		\def\arraystretch{1.3}
		\begin{tabular}{p{3.0cm} p{1.2cm} p{1.3cm} p{1.4cm} p{1.2cm} | p{2.0cm} p{2.0cm} p{2.0cm}}
			\hline
			& A: Wieber 2006a \cite{cite:wieber2006TrajectoryFree} & B: Wieber 2006b \cite{WIEBER2006559}& C: Stephens 2010 \cite{cite:stephens2010PushRecovery}& D: Urata 2011 \cite{cite:urata2011OnlineDecision} & E: Sagittal push w/o foot stepping & F: Sagittal push w/ foot stepping & G: Lateral push w/o foot stepping  \\
			\hline   
			Robot  & HRP-2 & Biped model & Sarcos Primus& HRP3L-JSK  & Valkyrie& Valkyrie& Valkyrie\\
			Robot height [$m$]& 1.539 & 1.425 & 1.575 & N/A &1.8&1.8&1.8\\
			CoM height [$m$] & N/A & N/A & N/A& 0.803 &1.1&1.1&1.1\\
			Mass [$kg$]& 58 & 40 & 92 & 53 & 137& 137& 137\\
			Force [$N$]& 1500 & 750.0 & N/A & 597& 600 & 2000 & 650\\
			Interval [$s$] & 0.025& 0.025& N/A& 0.05 & 0.12 & 0.12 & 0.12\\
			Impulse [$Ns$] & 37.5 & 18.8 & 42.0& 29.9 & 72.0 & 240.0 & 78.0\\
			Normalized impulse [$\frac{Ns}{kg}$] & 0.6 & 0.52 & 0.52 &0.6 & 0.57 & 1.73 & 0.56\\
			Stepping & No & No & Yes & Yes & No & Yes & No\\
			Simulated & Yes & Yes & No & No & Yes & Yes & Yes\\
			\hline 
		\end{tabular}
	\end{table*}
	
	\subsection{Realism of generated motions}
	\label{sec:reasonable}
	
	Despite learning is trained in a simulator, we emphasize realistic motions by enforcing joint angle, velocity, and torque limits, which are the same as on the real Valkyrie robot. Therefore, the learned motion could be applied on the real Valkyrie robot without violating physical constraints. Table \ref{tab:Peak_toruqes} compares the peak torques and velocities for different scenarios. The chosen scenarios require the largest joint torque for dropping and the largest joint velocities for taking multiple steps due to large pushes at the pelvis. All other presented test cases required less joint torque and velocity than the the ones presented in Table \ref{tab:Peak_toruqes}.
	\begin{table*}[t]
		\centering
		\caption{Peak torques and velocities for different scenarios.}
		\label{tab:Peak_toruqes}	  
		\def\arraystretch{1.3}
		\begin{tabular}{p{3.0cm} p{0.7cm} p{0.7cm} p{0.7cm} p{0.7cm} p{0.7cm} p{0.7cm} | p{0.7cm} p{0.7cm} p{0.7cm} p{0.7cm} p{0.7cm} p{0.7cm}}
			\hline
			& \multicolumn{6}{c}{Peak joint torque [$N$]} & \multicolumn{6}{c}{Peak joint velocity [$rad/s$]}\\
			\cline{2-13}
			& Torso pitch & Hip pitch & Hip Roll & Knee pitch & Ankle Pitch & Ankle Roll & Torso pitch & Hip pitch & Hip Roll & Knee pitch & Ankle Pitch & Ankle Roll\\
			\hline
			Joint limit & 150 & 350 & 350 & 350 & 205 & 205 & 9.00 &6.11 & 6.11 & 11.00 & 11.00 & 11.00 \\
			\hline
			Nominal standing & 68.4 & 39.5 & 57.1 & 122 & 44.9 & 46.2 & 0.0 & 0.0 & 0.0 & 0.0 & 0.0 & 0.0\\
			
			0.55m Drop  & 150 & 221 & 350 & 350 & 205 & 205 & 4.69 & 2.01 & 6.11 & 9.95 & 11.0 & 11.0\\
			78$Ns$ Pelvis (lateral)& 104 & 147 & 103 & 264 & 117 & 106 & 0.15 & 0.58 & 0.46 & 1.09 & 0.96 & 11.0\\
			240$Ns$ Pelvis (sagittal)& 150 & 187 & 350 & 350 & 205 & 205 & 6.49 & 1.47 & 6.11 & 5.69 & 11.0 & 11.0\\
			\hline 
		\end{tabular}
	\end{table*}
	
	\section{Conclusion}
	\label{sec:6}
	In this work we proposed a learning framework which is able to learn a versatile unified control policy via Deep Reinforcement Learning. We found that the policy is able to deal with different types of disturbances and has comparable performance to conventional controllers. The policy acquired is capable of functioning in unseen situations, demonstrating that it is generalizing well over tasks. Furthermore, the proposed learning framework is learning algorithm independent. We showed successful balance recovery with a policy trained with three of the state-of-the-art DRL algorithms: TRPO, PPO, and DDPG. The emerging motions for push recovery are similar to human motions demonstrating the ankle, hip, foot-tilting, and stepping strategy. We compared the learned policy against traditional push recovery controllers and found similar disturbance rejection capabilities.
	
	
	\bibliographystyle{IEEEtran}
	\bibliography{Humanoids2018}

\begin{thebibliography}{10}
\providecommand{\url}[1]{#1}
\csname url@rmstyle\endcsname
\providecommand{\newblock}{\relax}
\providecommand{\bibinfo}[2]{#2}
\providecommand\BIBentrySTDinterwordspacing{\spaceskip=0pt\relax}
\providecommand\BIBentryALTinterwordstretchfactor{4}
\providecommand\BIBentryALTinterwordspacing{\spaceskip=\fontdimen2\font plus
\BIBentryALTinterwordstretchfactor\fontdimen3\font minus
  \fontdimen4\font\relax}
\providecommand\BIBforeignlanguage[2]{{%
\expandafter\ifx\csname l@#1\endcsname\relax
\typeout{** WARNING: IEEEtran.bst: No hyphenation pattern has been}%
\typeout{** loaded for the language `#1'. Using the pattern for}%
\typeout{** the default language instead.}%
\else
\language=\csname l@#1\endcsname
\fi
#2}}

\bibitem{cite:yuan2018AnImprovedFormulation}
K.~Yuan and Z.~Li, ``An improved formulation for model predictive control of
  leggedrobots for gait planning and feedback control,'' in \emph{IROS}, 2018.

\bibitem{cite:hu2018ComparisonStudy}
W.~Hu, \emph{et~al.}, ``Comparison study of nonlinear optimization of step
  durations and foot placement for dynamic walking,'' in \emph{ICRA}, 2018.

\bibitem{cite:li2017RobustFoot}
Q.~Li, \emph{et~al.}, ``Robust foot placement control for dynamic walking using
  online parameter estimation,'' in \emph{Humanoids}, 2017.

\bibitem{cite:schulman2015TrustRegion}
J.~Schulman, \emph{et~al.}, ``Trust region policy optimization,'' in
  \emph{ICML}, 2015.

\bibitem{cite:lillicrap2016ContinuousControl}
T.~P. Lillicrap, \emph{et~al.}, ``Continuous control with deep reinforcement
  learning,'' \emph{arXiv:1509.02971}, 2015.

\bibitem{cite:schulman2017ProximalPolicy}
J.~Schulman, \emph{et~al.}, ``Proximal policy optimization algorithms,''
  \emph{arXiv:1707.06347}, 2017.

\bibitem{dallali2012global}
H.~Dallali, \emph{et~al.}, ``On global optimization of walking gaits for the
  compliant humanoid robot, coman using reinforcement learning,''
  \emph{Cybernetics and Information Technologies}, vol.~12, no.~3, pp. 39--52,
  2012.

\bibitem{cite:liu2018ImplementationOf}
C.~Liu, \emph{et~al.}, ``Implementation of deep deterministic policy gradients
  for controlling dynamic bipedal walking,'' in \emph{Conference on Biomimetic
  and Biohybrid Systems}, 2018.

\bibitem{cite:peng2017DeepLoco}
X.~B. Peng, \emph{et~al.}, ``{DeepLoco}: Dynamic locomotion skills using
  hierarchical deep reinforcement learning,'' \emph{Transactions on Graphics},
  2017.

\bibitem{cite:yu2018LearningSymmetric}
W.~Yu, \emph{et~al.}, ``Learning symmetry and low-energy locomotion,''
  \emph{arXiv:1801.08093}, 2018.

\bibitem{cite:stephens2007HumanoidPush}
B.~Stephens, ``Humanoid push recovery,'' in \emph{Humanoids}, 2007.

\bibitem{cite:li2017HumanoidBalancing}
Z.~Li, \emph{et~al.}, ``Humanoid balancing behavior featured by underactuated
  foot motion,'' \emph{T-RO}, 2017.

\bibitem{cite:wieber2006TrajectoryFree}
P.-B. Wieber, ``Trajectory free linear model predictive control for stable
  walking in the presence of strong perturbations,'' in \emph{Humanoids}, 2006.

\bibitem{cite:pratt2006CapturePoint}
J.~Pratt, \emph{et~al.}, ``Capture point: A step toward humanoid push
  recovery,'' in \emph{Humanoids}, 2006.

\bibitem{cite:komura2005SimulatingPathological}
T.~Komura, \emph{et~al.}, ``Simulating pathological gait using the enhanced
  linear inverted pendulum model,'' \emph{Transactions on Biomedical
  Engineering}, 2005.

\bibitem{cite:urata2011OnlineDecision}
J.~Urata, \emph{et~al.}, ``Online decision of foot placement using singular lq
  preview regulation,'' in \emph{Humanoids}, 2011.

\bibitem{cite:han2017FeedbackDesign}
W.~Han and R.~Tedrake, ``Feedback design for multi-contact push recovery via
  lmi approximation of the piecewise-affine quadratic regulator,'' in
  \emph{Humanoids}, 2017.

\bibitem{cite:marcucci2017ApproximateHybrid}
T.~Marcucci, \emph{et~al.}, ``Approximate hybrid model predictive control for
  multi-contact push recovery in complex environments,'' in \emph{Humanoids},
  Nov 2017.

\bibitem{cite:li2013Stabilizinghumanoids}
Z.~Li, \emph{et~al.}, ``Stabilizing humanoids on slopes using terrain
  inclination estimation,'' in \emph{IROS}, 2013.

\bibitem{cite:koolen2016BalanceControl}
T.~Koolen, \emph{et~al.}, ``Balance control using center of mass height
  variation: limitations imposed by unilateral contact,'' in \emph{Humanoids},
  2016.

\bibitem{cite:heess2017EmergenceOf}
N.~Heess, \emph{et~al.}, ``Emergence of locomotion behaviours in rich
  environments,'' \emph{arXiv:1707.02286}, 2017.

\bibitem{cite:ho2016GenerativeAdversarial}
J.~Ho and S.~Ermon, ``Generative adversarial imitation learning,'' in
  \emph{NIPS}, 2016.

\bibitem{cite:merel2017LearningHuman}
J.~Merel, \emph{et~al.}, ``Learning human behaviors from motion capture by
  adversarial imitation,'' \emph{arXiv:1707.02201}, 2017.

\bibitem{cite:peng2018Deepmimic}
X.~B. Peng, \emph{et~al.}, ``Deepmimic: Example-guided deep reinforcement
  learning of physics-based character skills,'' \emph{arXiv:1804.02717}, 2018.

\bibitem{cite:coumanspybullet}
E.~Coumans, \emph{et~al.}, ``Pybullet physics engine.''

\bibitem{cite:radford2015ValkyrieNasa}
N.~A. Radford, \emph{et~al.}, ``Valkyrie: {NASA}'s first bipedal humanoid
  robot,'' \emph{Journal of Field Robotics}, 2015.

\bibitem{cite:tensorflow2015-whitepaper}
\BIBentryALTinterwordspacing
M.~Abadi, \emph{et~al.}, ``{TensorFlow}: Large-scale machine learning on
  heterogeneous systems,'' 2015. [Online]. Available:
  \url{https://www.tensorflow.org/}
\BIBentrySTDinterwordspacing

\bibitem{cite:schulman2015HighDimensional}
J.~Schulman, \emph{et~al.}, ``High-dimensional continuous control using
  generalized advantage estimation,'' \emph{arXiv:1506.02438}, 2015.

\bibitem{cite:yang2017EmergenceOf}
C.~Yang, \emph{et~al.}, ``Emergence of human-comparable balancing behaviours by
  deep reinforcement learning,'' in \emph{Humanoids}, 2017.

\bibitem{cite:peng2016LearningLocomotion}
X.~B. Peng and M.~van~de Panne, ``Learning locomotion skills using deeprl: does
  the choice of action space matter?'' in \emph{SIGGRAPH}, 2017.

\bibitem{WIEBER2006559}
P.-B. Wieber and C.~Chevallereau, ``Online adaptation of reference trajectories
  for the control of walking systems,'' \emph{Robotics and Autonomous Systems},
  vol.~54, no.~7, pp. 559 -- 566, 2006.

\bibitem{cite:stephens2010PushRecovery}
B.~J. Stephens and C.~G. Atkeson, ``Push recovery by stepping for humanoid
  robots with force controlled joints,'' in \emph{Humanoids}, 2010.

\end{thebibliography}
	\newpage
	
	\section{Supplementary Material}
	\subsection{Hyperparameters for Deep Reinforcement Learning}
	\subsubsection{TRPO}
	
Policy Network: (100, tanh, 50, tanh, 25, tanh, Linear) + Standard Deviation variable

Value Network (100, ReLU, 50, ReLU, 25, ReLU, linear)

Timesteps per batch = 4096

max KL divergence = 0.01

Conjugate gradient iterations = 10

Conjugate gradient damping = 0.1

Value function iterations = 10

Value function batch size = 256

Value function step size = 3e − 4

Entropy coefficient = 0.0

Discount $\gamma$ = 0.95, GAE $\lambda$ = 0.95
	
	\subsubsection{PPO}

Policy Network: (100, tanh, 50, tanh, 25, tanh, Linear) + Standard Deviation variable

Value Network (100, ReLU, 50, ReLU, 25, ReLU, linear)

Timesteps per batch = 4096

Clip parameter = 0.2

Entropy coefficient = 0.0

Optimizer epochs per iteration = 10

Optimizer step size = 3e − 4

Optimizer batch size = 256

Discount $\gamma$ = 0.95, GAE $\lambda$ = 0.95

%

\end{document}